\documentclass[10pt,twocolumn,letterpaper]{article}

 \usepackage{iccv}              

\definecolor{iccvblue}{rgb}{0.21,0.49,0.74}
\usepackage[pagebackref,breaklinks,colorlinks,allcolors=iccvblue]{hyperref}
\usepackage{multirow, array}
\usepackage{makecell}
\usepackage{booktabs} 
\newcolumntype{C}[1]{>{\centering\arraybackslash}p{#1}}
\def\paperID{1} 
\def\confName{ICCV}
\def\confYear{2025}

\title{Preview WB-DH: Towards Whole Body Digital Human Bench for the Generation of Whole-body Talking Avatar Videos}

\author{Chaoyi Wang$^1$, Yifan Yang$^2$, Jun Pei$^1$, Lijie Xia$^1$, Jianpo Liu$^1$, Xiaobing Yuan$^{1*}$, Xinhan Di$^{3*}$\\
$^1$Shanghai Institute of Microsystem and Information Technology, CAS, China\\
$^2$Shanghai Jiao Tong University, China \quad $^3$Independent Researcher, China\\
{\tt\small \{chaoyiwang,peijun,xialj,liujp,sinoiot\}@mail.sim.ac.cn, v1o2058@sjtu.edu.cn, deepearthgo@gmail.com}
}
\begin{document}
\maketitle
\begin{abstract}
Creating realistic, fully animatable whole-body avatars from a single portrait is challenging due to limitations in capturing subtle expressions, body movements, and dynamic backgrounds. Current evaluation datasets and metrics fall short in addressing these complexities. To bridge this gap, we introduce the Whole-Body Benchmark Dataset (WB-DH), an open-source, multi-modal benchmark designed for evaluating whole-body animatable avatar generation. Key features include: (1) detailed multi-modal annotations for fine-grained guidance, (2) a versatile evaluation framework, and (3) public access to the dataset and tools at \href{https://github.com/deepreasonings/WholeBodyBenchmark}{https://github.com/deepreasonings/WholeBodyBenchmark}.
\end{abstract}    
\section{Introduction}
\label{sec:intro}
Diffusion-based video generation has emerged as a powerful approach for synthesizing realistic videos~\cite{singer2022make}. Researchers are exploring cascaded and latent-space diffusion pipelines to mitigate these issues~\cite{li2023videogen,zhang2023i2vgen,wang2024lavie}. In the context of digital humans~\cite{prajwal2020lip,min2022styletalker}, current diffusion models' synthesis of plausible full-body motion and appearance is beyond practice~\cite{cui2024hallo2, cui2024hallo3,ji2024sonic,guan2023stylesync,meng2024echomimicv2,tu2024stableanimator}, and they rarely incorporate audio for speech synchronization or full-body gesture generation. This gap motivates the need for integrated systems that handle both high-quality full-body motion and speech generation, which is the focus of the Whole Body Digital Human Bench.

However, achieving fully natural, talking whole-body avatars remains a significant challenge. Yet even these advanced talking head models typically concentrate on the face and half-body region ~\cite{cui2024hallo2, cui2024hallo3,ji2024sonic,guan2023stylesync} and do not synthesize full-body motions or gestures tied to speech ~\cite{meng2024echomimicv2,tu2024stableanimator}. This limitation highlights the importance of the Whole-Body Digital Human Bench, designed to evaluate and advance methods that unify speech-driven facial animation with coherent whole-body generation. 
\begin{figure*}[!ht] 
    \centering
    \includegraphics[width=0.85\textwidth]{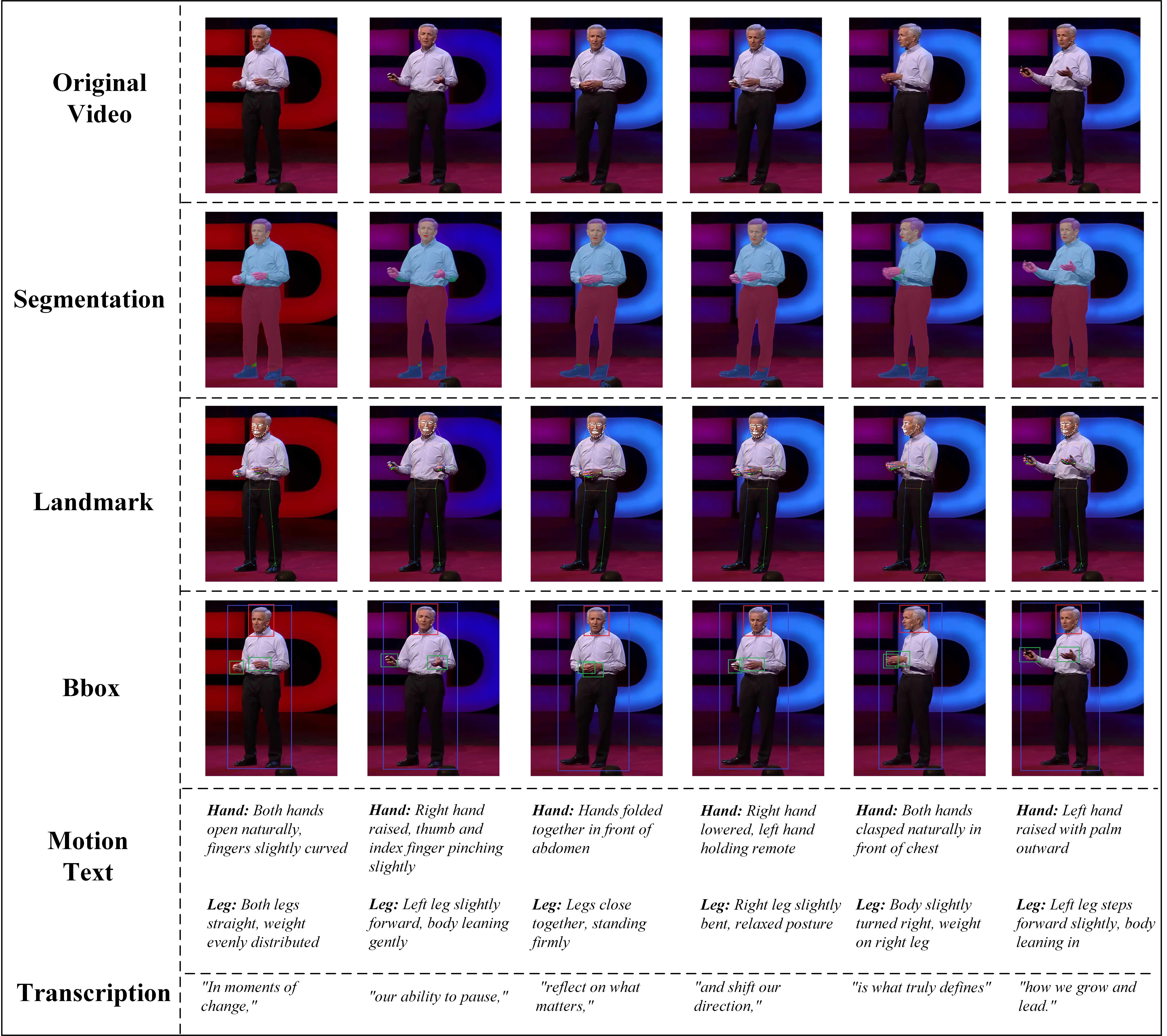}
    \caption{Illustration of our dataset annotations. Each column represents a key frame. From top to bottom: the original frame, body segmentation, landmark, bounding box annotations for key regions (hands, legs, whole body), motion text describing pose semantics, and corresponding speech transcription. The dataset captures fine-grained multi-modal alignment between body posture, hand gestures, leg stance, and spoken language across time.}
    \label{fig:1}
\end{figure*}

\begin{figure*}[!ht] 
    \centering
    \includegraphics[width=1.0\textwidth]{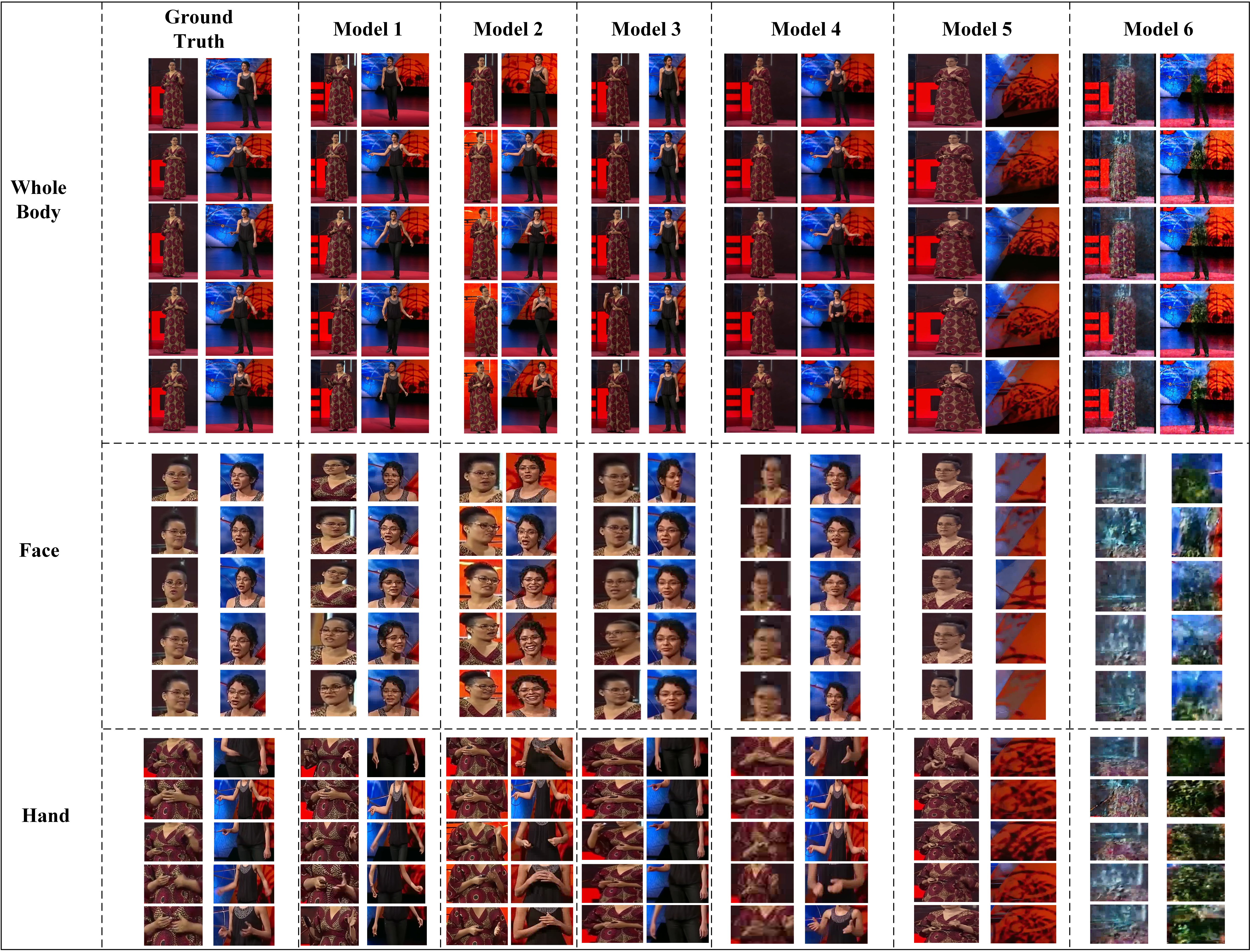}
    \caption{Visual comparison across full body, face, and hand regions. Model 1(Wan~\cite{wan2025}),2(OpenS~\cite{opensora2}),3(Hun~\cite{kong2024hunyuanvideo}), 4(ecv2/w~\cite{meng2024echomimicv2}) show good performance in structure, motion, and identity preservation. Model 5(ecv2/wo~\cite{meng2024echomimicv2}),6(Ha3/wo~\cite{cui2024hallo3}) exhibits noticeable artifacts and degradation across all regions.}
    \label{fig:2}
\end{figure*}

\section{Related Work}
\label{sec:relatedwork}
\subsection{Diffusion-Based Video Generation}
A number of diffusion-based frameworks have pushed the boundaries of video synthesis. Initial work extended image diffusion models into the temporal domain~\cite{ho2022video}. Building on this foundation, later research incorporated conditioning information for high-quality generated content ~\cite{li2023videogen,singer2022make}. Other work focuses on efficiency and scale~\cite{kong2024hunyuanvideo,qiu2025skyreels,xu2024easyanimate,yang2024cogvideox,luma_ray2,google2025gemini,guo2023animatediff,huang2025step}, which is evidenced by Stable Video Diffusion on large datasets~\cite{blattmann2023stable}. However, current video generation models do not inherently ensure that a character’s speech aligns with their whole-body visuals. This gap underscores the need for benchmarks that integrate audio with synthesized whole-body video generation.

\subsection{Audio-Driven Talking Head Generation}
Recently, accurate lip synchronization is applied via a specialized lip-sync discriminator to guide a GAN-based face generator~\cite{prajwal2020lip,zhou2021pose,min2022styletalker}. Besides, diffusion models have made significant inroads in this field~\cite{stypulkowski2024diffused,chopin2025dimitra}. However, as with diffusion video models, existing talking head techniques are typically limited to the head, shoulders, and upper-body region~\cite{jiang2024loopy,li2024latentsync,cui2024hallo2, cui2024hallo3,ji2024sonic,guan2023stylesync}. This disconnect emphasizes the need for a comprehensive evaluation platform — the Whole-Body Digital Human Bench — to assess and promote methods that integrate audio-driven facial animation with full-body video generation.

\renewcommand{\arraystretch}{1.0} 
\setlength{\tabcolsep}{0pt}        
\begin{table*}[!ht]
\centering
\caption{Evaluation results of eight models across three regions and twelve metrics. SC: Subject Consistency, BC: Background Consistency, MS: Motion Smoothness, DD: Dynamic Degree, AQ: Aesthetic Quality, IQ: Imaging Quality, FID: Fréchet Inception Distance, FVD: Fréchet Video Distance, SSIM: Structural Similarity Index Measure, PSNR: Peak Signal-to-Noise Ratio, E-FID: Enhanced Fréchet Inception Distance, CSIM: Cosine Similarity.}
\label{tab:score_table}
\footnotesize
\begin{tabular}{C{0.1\linewidth}|C{0.1\linewidth}|C{0.08\linewidth}|*{8}{C{0.09\linewidth}}}
\toprule
\textbf{Region} & \textbf{Metric} & \textbf{GT} & \textbf{Step}~\cite{huang2025step} & \textbf{Hun}~\cite{kong2024hunyuanvideo}& \textbf{Wan}~\cite{wan2025} & \textbf{OpenS}~\cite{opensora2} & \textbf{Ha3/w}~\cite{cui2024hallo3} & \textbf{Ha3/wo}~\cite{cui2024hallo3} & \textbf{ecv2/w}~\cite{meng2024echomimicv2} & \textbf{ecv2/wo}~\cite{meng2024echomimicv2} \\
\midrule
\multirow{14}{*}{Whole Body}
& SC & 100\% & 93.45\% & 92.54\% & \textbf{96.83\%} & 97.12\% & 20.20\% & 10.74\% & 29.16\% & 7.08\% \\
& BC & 100\% & 94.50\% & 94.30\% & \textbf{96.31\%} & 98.00\% & 19.91\% & 10.93\%  & 31.25\% & 6.49\% \\
& MS & 100\% & 98.71\% & 98.50\% & \textbf{99.69\%} & 90.69\% & 20.60\% & 11.37\% & 29.75\% & 7.14\% \\
& DD & 100\% & 67.39\% & \textbf{69.34\%} & 59.89\% & 60.43\% & 4.65\% & 4.00\% & 9.81\% & 0.09\%\\
& AQ & 100\% & \textbf{48.32\%} & 43.17\% & 46.73\% & 32.88\% & 10.01\% & 3.84\% & 14.27\% & 2.52\%\\
& IQ & 100\% & 61.99\% & \textbf{63.54\%} & 54.67\% & 42.57\% & 9.88\% & 5.09\% & 16.53\% & 3.29\% \\
\cmidrule{2-11}
& FID & 0.00 & 172.47 & 208.75 & \textbf{92.76} & 121.53 & 491.43 & 568.68 & 508.19 & 572.14 \\
& FVD & 0.00 & 1356.49 & 1567.94 & \textbf{750.51} & 896.94 & 2470.14 & 4366.39 & 2519.77 & 3791.25 \\
& SSIM & 1.00 & 0.573 & 0.543 & \textbf{0.660} & 0.623 & 0.209 & 0.016 & 0.216 & 0.028 \\
& PSNR & $\infty$ & 17.74 & 16.06 & \textbf{20.16} & 19.48 & 6.24 & 0.83 & 6.17 & 1.29 \\
& E-FID & 0.00 & \textbf{187.58} & 216.91 & 196.39 & 218.78 & 484.08 & 634.60 & 492.17 & 712.91 \\
& CSIM & 1.00 & 0.760 & 0.811 & \textbf{0.896} & 0.876 & 0.303 & 0.053 & 0.299 & 0.092 \\
\midrule
\multirow{14}{*}{Face}
& SC & 100\% & 65.06\% & 55.07\% & \textbf{66.51\%} & 52.34\% & 14.92\% & 12.23\% & 14.79\% & 8.29\% \\
& BC & 100\% & \textbf{74.04\%} & 68.05\% & 72.26\% & 63.55\% & 15.73\% & 15.15\% & 18.16\% & 9.27\% \\ 
& MS & 100\% & 63.12\% & 56.78\% & \textbf{63.40\%} & 49.74\% & 14.66\% & 12.15\% & 12.73\%  & 8.16\% \\
& DD & 100\% & 14.91\% & 4.15\%  & \textbf{25.12\%} & 7.69\% & 7.21\% & 4.55\% & 2.42\% & 0.21\%  \\
& AQ & 100\% & \textbf{28.18\%} & 21.26\% & 25.09\% & 19.76\% & 5.79\% & 2.99\% & 5.71\% & 5.69\% \\
& IQ & 100\% & \textbf{39.40\%} & 34.19\% & 33.68\% & 30.90\% & 6.98\% & 4.60\% & 8.49\% & 6.29\%  \\
\cmidrule{2-11}
& FID & 0.00 & 300.98 & 291.21 & \textbf{254.03} & 275.50 & 425.82 & 720.22 & 437.27 & 801.27 \\
& FVD & 0.00 & 1811.92 & 1351.71 & 1591.22 & \textbf{1221.85} & 2794.85 & 3255.19 & 2651.92 & 3047.27 \\
& SSIM & 1.00 & 0.372 & 0.322 & \textbf{0.417} & 0.388 & 0.127 & 0.010 & 0.139 & 0.027 \\
& PSNR & $\infty$ & 13.54 & 14.47 & \textbf{16.99} & 16.29 & 6.75 & 0.89 & 6.88 & 1.249 \\
& E-FID & 0.00 & 339.18 & 347.24 & \textbf{273.97} & 359.85 & 451.76 & 492.67 & 439.92 & 513.58 \\
& CSIM & 1.00 & 0.736 & 0.787 & \textbf{0.866} & 0.863 & 0.385 & 0.065 & 0.401 & 0.062 \\
\midrule
\multirow{14}{*}{Hand}
& SC & 100\% & \textbf{66.83\%} & 48.15\% & 51.27\% & 49.16\% & 13.65\% & 2.13\% & 14.19\% & 3.58\% \\
& BC & 100\% & \textbf{77.98\%} & 64.98\% & 57.45\% & 32.45\% & 14.76\% & 2.56\% & 18.21\% & 4.27\% \\
& MS & 100\% & 46.36\% & \textbf{49.65\%} & 42.49\% & 43.30\% & 13.25\% & 0.49\% & 9.27\% & 2.79\% \\
& DD & 100\% & 48.14\% & \textbf{52.85\%} & 41.71\% & 43.89\% & 13.05\% & 0.91\% & 9.52\% & 1.84\% \\
& AQ & 100\% & \textbf{24.22\%} & 19.02\% & 18.38\% & 16.40\% & 5.08\% & 0.39\% & 5.08\% & 2.74\%  \\ 
& IQ & 100\% & \textbf{38.40\%} & 27.69\% & 23.84\% & 21.50\% & 6.51\% & 0.89\% & 10.24\% & 1.94\% \\
\cmidrule{2-11}
& FID & 0.00 & 327.10 & 301.94 & 331.51 & \textbf{278.95} & 574.40 & 921.16 & 588.72 & 975.92 \\
& FVD & 0.00 & 1597.65 & 1734.98 & 1561.58 & \textbf{1534.50} & 7750.48 & 10671.79 & 8958.27 & 9472.28 \\
& SSIM & 1.00 & 0.316 & 0.262 & \textbf{0.367} & 0.342 & 0.098 & 0.008 & 0.106 & 0.019 \\
& PSNR & $\infty$ & 12.98 & 11.45 & 13.09 & \textbf{15.50} & 6.13 & 0.55 & 7.272 & 0.492 \\
& E-FID & 0.00 & 277.47 & 255.53 & \textbf{240.48} & 264.98 & 445.59 & 713.46 & 525.18 & 741.03 \\
& CSIM & 1.00 & 0.681 & 0.737 & \textbf{0.796} & 0.794 & 0.366 & 0.041 & 0.427 & 0.152 \\
\bottomrule
\end{tabular}
\end{table*}

\section{Method}
In this section, we first introduce our large-scale dataset, followed by a detailed explanation of our region-specific evaluation strategy. Finally, we present the quantitative results across multiple state-of-the-art video generation models and evaluation dimensions.

\subsection{Dataset Construction}
To facilitate large-scale learning of speech-driven digital human generation, we curate a comprehensive dataset that includes over \textbf{10,000 unique identities}, each appearing in approximately \textbf{200 different scene configurations}, totaling \textbf{2M samples(video clips)}, the annotation for each sample is demonstrated in Figure~\ref{fig:1}.\textbf{20K samples} are sampled for the following test.

\subsection{Video Generation Sub Evaluation Protocol}
To systematically assess the perceptual quality and temporal stability of the generated whole-body animatable avatar, six automatic, reference-free evaluation metrics tailored for video generation tasks are applied. 

The six metrics are briefly defined as the follows:
\begin{itemize}
    \item \textbf{Subject Consistency (SC)}: Temporal consistency of subject appearance, measured using DINO feature similarity~\cite{caron2021emerging}.
    \item \textbf{Background Consistency (BC)}: Stability of background scenes, measured with CLIP feature similarity~\cite{radford2021learning}.
    \item \textbf{Motion Smoothness (MS)}: Smoothness and physical plausibility of motion, computed using optical flow continuity~\cite{li2023amt}.
    \item \textbf{Dynamic Degree (DD)}: Richness of motion, estimated via average optical flow magnitude~\cite{teed2020raft}.
    \item \textbf{Aesthetic Quality (AQ)}: Perceptual attractiveness scored using the LAION aesthetic predictor~\cite{laion_aesthetic_predictor}.
    \item \textbf{Imaging Quality (IQ)}: Sharpness and clarity evaluated using MUSIQ~\cite{ke2021musiq} trained on the SPAQ dataset.
\end{itemize}
These metrics together provide a holistic view of both visual realism and temporal coherence in generated sequences. Each model under study is evaluated using these criteria.
\subsection{Co-Speech Sub Evaluation Protocol}
To complement the region-level evaluation of the whole-body generated avatars, we adopt six standard metrics to assess frame quality, temporal coherence, and identity preservation~\cite{cui2024hallo2, cui2024hallo3,ji2024sonic,guan2023stylesync}. 
\begin{itemize} \item \textbf{Fréchet Inception Distance (FID)}: Measures distributional similarity between generated and real images using deep features from an Inception network. Lower is better~\cite{qin2025versatile}.
\item \textbf{Fréchet Video Distance (FVD)}: Extends FID to videos using temporal coherence features to account for both appearance and temporal dynamics. Strongly correlates with human perception~\cite{unterthiner2018towards}.
\item \textbf{Structural Similarity (SSIM)}: Estimates frame-wise perceptual similarity by comparing structural and luminance information between generated and ground-truth frames~\cite{wang2004image}.
\item \textbf{Peak Signal-to-Noise Ratio (PSNR)}: Evaluates pixel-wise fidelity through mean squared error. Sensitive to distortions but less aligned with human perception~\cite{huynh2012accuracy}.
\item \textbf{Extended FID (E-FID)}: An improved variant of FID for portrait video generation, focusing on visual realism and authenticity in identity-sensitive tasks~\cite{qin2025versatile}.
\item \textbf{Cosine Similarity Identity Metric (CSIM)}: Measures identity preservation by computing the cosine similarity between identity embeddings from reference and generated frames~\cite{qin2025versatile}.
\end{itemize}
These metrics provide complementary insights into generation quality and are used in the proposed dataset to ensure a comprehensive evaluation.
\section{Initial Evaluation}
We evaluate two groups of state-of-the-art (SOTA) models: (1) open-source video generation models~\cite{kong2024hunyuanvideo, huang2025step, wan2025, opensora2}, and (2) open-source talking avatar models~\cite{meng2024echomimicv2, cui2024hallo3}. Table~\ref{tab:score_table} presents the evaluation metrics, while Figure~\ref{fig:2} illustrates initial region-specific generation results for the face, hands, and full body, respectively. Given the first frame of a video clip as input, video generation models are guided by a text description of the intended motion (Figure~\ref{fig:1}). In contrast, talking avatar models are driven by audio input, with or without additional body pose guidance (Figure~\ref{fig:1}). We refer to these variants as Ha3/w and Ha3/wo~\cite{cui2024hallo3}, representing models with and without pose guidance, respectively. To capture region-level differences in generative quality, we conduct independent evaluations for three spatial regions: full body, face, and hands.
\section{Conclusion}
We propose a large-scale dataset and a comprehensive set of evaluation metrics for whole-body talking avatar generation. In the initial assessment, a range of open-source state-of-the-art models are evaluated. We are currently expanding the dataset to incorporate more complex motions during speech and extending the evaluation metrics accordingly. In future work, we plan to include closed-source models in the evaluation. Additionally, we are developing a formal Version $1$ of the dataset to support joint audio-video generation, extending the video duration from $10$ seconds to $60$ seconds.

{
    \small
    \bibliographystyle{ieeenat_fullname}
    \bibliography{main}

\begin{thebibliography}{39}
\providecommand{\natexlab}[1]{#1}
\providecommand{\url}[1]{\texttt{#1}}
\expandafter\ifx\csname urlstyle\endcsname\relax
  \providecommand{\doi}[1]{doi: #1}\else
  \providecommand{\doi}{doi: \begingroup \urlstyle{rm}\Url}\fi

\bibitem[Blattmann et~al.(2023)Blattmann, Dockhorn, Kulal, Mendelevitch, Kilian, Lorenz, Levi, English, Voleti, Letts, et~al.]{blattmann2023stable}
Andreas Blattmann, Tim Dockhorn, Sumith Kulal, Daniel Mendelevitch, Maciej Kilian, Dominik Lorenz, Yam Levi, Zion English, Vikram Voleti, Adam Letts, et~al.
\newblock Stable video diffusion: Scaling latent video diffusion models to large datasets.
\newblock \emph{arXiv preprint arXiv:2311.15127}, 2023.

\bibitem[Caron et~al.(2021)Caron, Touvron, Misra, J{\'e}gou, Mairal, Bojanowski, and Joulin]{caron2021emerging}
Mathilde Caron, Hugo Touvron, Ishan Misra, Herv{\'e} J{\'e}gou, Julien Mairal, Piotr Bojanowski, and Armand Joulin.
\newblock Emerging properties in self-supervised vision transformers.
\newblock In \emph{Proceedings of the IEEE/CVF international conference on computer vision}, pages 9650--9660, 2021.

\bibitem[Chopin et~al.(2025)Chopin, Dhamija, Balaji, Wang, and Dantcheva]{chopin2025dimitra}
Baptiste Chopin, Tashvik Dhamija, Pranav Balaji, Yaohui Wang, and Antitza Dantcheva.
\newblock Dimitra: Audio-driven diffusion model for expressive talking head generation.
\newblock \emph{arXiv preprint arXiv:2502.17198}, 2025.

\bibitem[Cui et~al.(2024{\natexlab{a}})Cui, Li, Yao, Zhu, Shang, Cheng, Zhou, Zhu, and Wang]{cui2024hallo2}
Jiahao Cui, Hui Li, Yao Yao, Hao Zhu, Hanlin Shang, Kaihui Cheng, Hang Zhou, Siyu Zhu, and Jingdong Wang.
\newblock Hallo2: Long-duration and high-resolution audio-driven portrait image animation.
\newblock \emph{arXiv preprint arXiv:2410.07718}, 2024{\natexlab{a}}.

\bibitem[Cui et~al.(2024{\natexlab{b}})Cui, Li, Zhan, Shang, Cheng, Ma, Mu, Zhou, Wang, and Zhu]{cui2024hallo3}
Jiahao Cui, Hui Li, Yun Zhan, Hanlin Shang, Kaihui Cheng, Yuqi Ma, Shan Mu, Hang Zhou, Jingdong Wang, and Siyu Zhu.
\newblock Hallo3: Highly dynamic and realistic portrait image animation with diffusion transformer networks.
\newblock \emph{arXiv preprint arXiv:2412.00733}, 2024{\natexlab{b}}.

\bibitem[{Google DeepMind}(2025)]{google2025gemini}
{Google DeepMind}.
\newblock Gemini 2.5 pro: Our most intelligent ai model, 2025.
\newblock Accessed: 2025-04-25.

\bibitem[Guan et~al.(2023)Guan, Zhang, Zhou, Hu, Wang, He, Feng, Liu, Ding, Liu, et~al.]{guan2023stylesync}
Jiazhi Guan, Zhanwang Zhang, Hang Zhou, Tianshu Hu, Kaisiyuan Wang, Dongliang He, Haocheng Feng, Jingtuo Liu, Errui Ding, Ziwei Liu, et~al.
\newblock Stylesync: High-fidelity generalized and personalized lip sync in style-based generator.
\newblock In \emph{Proceedings of the IEEE/CVF Conference on Computer Vision and Pattern Recognition}, pages 1505--1515, 2023.

\bibitem[Guo et~al.(2023)Guo, Yang, Rao, Liang, Wang, Qiao, Agrawala, Lin, and Dai]{guo2023animatediff}
Yuwei Guo, Ceyuan Yang, Anyi Rao, Zhengyang Liang, Yaohui Wang, Yu Qiao, Maneesh Agrawala, Dahua Lin, and Bo Dai.
\newblock Animatediff: Animate your personalized text-to-image diffusion models without specific tuning.
\newblock \emph{arXiv preprint arXiv:2307.04725}, 2023.

\bibitem[Ho et~al.(2022)Ho, Salimans, Gritsenko, Chan, Norouzi, and Fleet]{ho2022video}
Jonathan Ho, Tim Salimans, Alexey Gritsenko, William Chan, Mohammad Norouzi, and David~J Fleet.
\newblock Video diffusion models.
\newblock \emph{Advances in Neural Information Processing Systems}, 35:\penalty0 8633--8646, 2022.

\bibitem[Huang et~al.(2025)Huang, Ma, Duan, Chen, Wan, Ming, Wang, Wang, Lu, Li, et~al.]{huang2025step}
Haoyang Huang, Guoqing Ma, Nan Duan, Xing Chen, Changyi Wan, Ranchen Ming, Tianyu Wang, Bo Wang, Zhiying Lu, Aojie Li, et~al.
\newblock Step-video-ti2v technical report: A state-of-the-art text-driven image-to-video generation model.
\newblock \emph{arXiv preprint arXiv:2503.11251}, 2025.

\bibitem[Huynh-Thu and Ghanbari(2012)]{huynh2012accuracy}
Quan Huynh-Thu and Mohammed Ghanbari.
\newblock The accuracy of psnr in predicting video quality for different video scenes and frame rates.
\newblock \emph{Telecommunication systems}, 49:\penalty0 35--48, 2012.

\bibitem[Ji et~al.(2024)Ji, Hu, Xu, Zhu, Lin, He, Zhang, Luo, Chen, Lin, et~al.]{ji2024sonic}
Xiaozhong Ji, Xiaobin Hu, Zhihong Xu, Junwei Zhu, Chuming Lin, Qingdong He, Jiangning Zhang, Donghao Luo, Yi Chen, Qin Lin, et~al.
\newblock Sonic: Shifting focus to global audio perception in portrait animation.
\newblock \emph{arXiv preprint arXiv:2411.16331}, 2024.

\bibitem[Jiang et~al.(2024)Jiang, Liang, Yang, Lin, Zhong, and Zheng]{jiang2024loopy}
Jianwen Jiang, Chao Liang, Jiaqi Yang, Gaojie Lin, Tianyun Zhong, and Yanbo Zheng.
\newblock Loopy: Taming audio-driven portrait avatar with long-term motion dependency.
\newblock \emph{arXiv preprint arXiv:2409.02634}, 2024.

\bibitem[Ke et~al.(2021)Ke, Wang, Wang, Milanfar, and Yang]{ke2021musiq}
Junjie Ke, Qifei Wang, Yilin Wang, Peyman Milanfar, and Feng Yang.
\newblock Musiq: Multi-scale image quality transformer.
\newblock In \emph{Proceedings of the IEEE/CVF international conference on computer vision}, pages 5148--5157, 2021.

\bibitem[Kong et~al.(2024)Kong, Tian, Zhang, Min, Dai, Zhou, Xiong, Li, Wu, Zhang, et~al.]{kong2024hunyuanvideo}
Weijie Kong, Qi Tian, Zijian Zhang, Rox Min, Zuozhuo Dai, Jin Zhou, Jiangfeng Xiong, Xin Li, Bo Wu, Jianwei Zhang, et~al.
\newblock Hunyuanvideo: A systematic framework for large video generative models.
\newblock \emph{arXiv preprint arXiv:2412.03603}, 2024.

\bibitem[Labs()]{luma_ray2}
Luma Labs.
\newblock Luma ray 2 video model.
\newblock \url{https://lumalabs.ai/ray}.
\newblock Accessed: 2025-04-25.

\bibitem[{LAION-AI}(2022)]{laion_aesthetic_predictor}
{LAION-AI}.
\newblock Laion aesthetic predictor.
\newblock \url{https://github.com/LAION-AI/aesthetic-predictor}, 2022.
\newblock Accessed: 2025-04-21.

\bibitem[Li et~al.(2024)Li, Zhang, Xu, Xie, Feng, Peng, and Xing]{li2024latentsync}
Chunyu Li, Chao Zhang, Weikai Xu, Jinghui Xie, Weiguo Feng, Bingyue Peng, and Weiwei Xing.
\newblock Latentsync: Audio conditioned latent diffusion models for lip sync.
\newblock \emph{arXiv preprint arXiv:2412.09262}, 2024.

\bibitem[Li et~al.(2023{\natexlab{a}})Li, Chu, Wu, Yuan, Liu, Zhang, Li, Feng, Ding, and Wang]{li2023videogen}
Xin Li, Wenqing Chu, Ye Wu, Weihang Yuan, Fanglong Liu, Qi Zhang, Fu Li, Haocheng Feng, Errui Ding, and Jingdong Wang.
\newblock Videogen: A reference-guided latent diffusion approach for high definition text-to-video generation.
\newblock \emph{arXiv preprint arXiv:2309.00398}, 2023{\natexlab{a}}.

\bibitem[Li et~al.(2023{\natexlab{b}})Li, Zhu, Han, Hou, Guo, and Cheng]{li2023amt}
Zhen Li, Zuo-Liang Zhu, Ling-Hao Han, Qibin Hou, Chun-Le Guo, and Ming-Ming Cheng.
\newblock Amt: All-pairs multi-field transforms for efficient frame interpolation.
\newblock In \emph{Proceedings of the IEEE/CVF Conference on Computer Vision and Pattern Recognition}, pages 9801--9810, 2023{\natexlab{b}}.

\bibitem[Meng et~al.(2024)Meng, Zhang, Li, and Ma]{meng2024echomimicv2}
Rang Meng, Xingyu Zhang, Yuming Li, and Chenguang Ma.
\newblock Echomimicv2: Towards striking, simplified, and semi-body human animation.
\newblock \emph{arXiv preprint arXiv:2411.10061}, 2024.

\bibitem[Min et~al.(2022)Min, Song, Ko, and Hwang]{min2022styletalker}
Dongchan Min, Minyoung Song, Eunji Ko, and Sung~Ju Hwang.
\newblock Styletalker: One-shot style-based audio-driven talking head video generation.
\newblock \emph{arXiv preprint arXiv:2208.10922}, 2022.

\bibitem[Peng et~al.(2025)Peng, Zheng, Shen, Young, Guo, Wang, Xu, Liu, Jiang, Li, Wang, Ye, Ren, Ma, Liang, Lian, Wu, Zhong, Li, Gong, Lei, Cheng, Zhang, Li, Zhang, Hu, Huang, Wang, Zhao, Wang, Wei, and You]{opensora2}
Xiangyu Peng, Zangwei Zheng, Chenhui Shen, Tom Young, Xinying Guo, Binluo Wang, Hang Xu, Hongxin Liu, Mingyan Jiang, Wenjun Li, Yuhui Wang, Anbang Ye, Gang Ren, Qianran Ma, Wanying Liang, Xiang Lian, Xiwen Wu, Yuting Zhong, Zhuangyan Li, Chaoyu Gong, Guojun Lei, Leijun Cheng, Limin Zhang, Minghao Li, Ruijie Zhang, Silan Hu, Shijie Huang, Xiaokang Wang, Yuanheng Zhao, Yuqi Wang, Ziang Wei, and Yang You.
\newblock Open-sora 2.0: Training a commercial-level video generation model in \$200k.
\newblock \emph{arXiv preprint arXiv:2503.09642}, 2025.

\bibitem[Prajwal et~al.(2020)Prajwal, Mukhopadhyay, Namboodiri, and Jawahar]{prajwal2020lip}
KR Prajwal, Rudrabha Mukhopadhyay, Vinay~P Namboodiri, and CV Jawahar.
\newblock A lip sync expert is all you need for speech to lip generation in the wild.
\newblock In \emph{Proceedings of the 28th ACM international conference on multimedia}, pages 484--492, 2020.

\bibitem[Qin et~al.(2025)Qin, Zheng, Wang, Li, Zhu, Yang, Yang, and Wang]{qin2025versatile}
Zheng Qin, Ruobing Zheng, Yabing Wang, Tianqi Li, Zixin Zhu, Minghui Yang, Ming Yang, and Le Wang.
\newblock Versatile multimodal controls for whole-body talking human animation.
\newblock \emph{arXiv preprint arXiv:2503.08714}, 2025.

\bibitem[Qiu et~al.(2025)Qiu, Fei, Wang, Bai, Yu, Fan, Chen, and Wen]{qiu2025skyreels}
Di Qiu, Zhengcong Fei, Rui Wang, Jialin Bai, Changqian Yu, Mingyuan Fan, Guibin Chen, and Xiang Wen.
\newblock Skyreels-a1: Expressive portrait animation in video diffusion transformers.
\newblock \emph{arXiv preprint arXiv:2502.10841}, 2025.

\bibitem[Radford et~al.(2021)Radford, Kim, Hallacy, Ramesh, Goh, Agarwal, Sastry, Askell, Mishkin, Clark, et~al.]{radford2021learning}
Alec Radford, Jong~Wook Kim, Chris Hallacy, Aditya Ramesh, Gabriel Goh, Sandhini Agarwal, Girish Sastry, Amanda Askell, Pamela Mishkin, Jack Clark, et~al.
\newblock Learning transferable visual models from natural language supervision.
\newblock In \emph{International conference on machine learning}, pages 8748--8763. PmLR, 2021.

\bibitem[Singer et~al.(2022)Singer, Polyak, Hayes, Yin, An, Zhang, Hu, Yang, Ashual, Gafni, et~al.]{singer2022make}
Uriel Singer, Adam Polyak, Thomas Hayes, Xi Yin, Jie An, Songyang Zhang, Qiyuan Hu, Harry Yang, Oron Ashual, Oran Gafni, et~al.
\newblock Make-a-video: Text-to-video generation without text-video data.
\newblock \emph{arXiv preprint arXiv:2209.14792}, 2022.

\bibitem[Stypu{\l}kowski et~al.(2024)Stypu{\l}kowski, Vougioukas, He, Zikeba, Petridis, and Pantic]{stypulkowski2024diffused}
Micha{\l} Stypu{\l}kowski, Konstantinos Vougioukas, Sen He, Maciej Zikeba, Stavros Petridis, and Maja Pantic.
\newblock Diffused heads: Diffusion models beat gans on talking-face generation.
\newblock In \emph{Proceedings of the IEEE/CVF Winter Conference on Applications of Computer Vision}, pages 5091--5100, 2024.

\bibitem[Teed and Deng(2020)]{teed2020raft}
Zachary Teed and Jia Deng.
\newblock Raft: Recurrent all-pairs field transforms for optical flow.
\newblock In \emph{Computer Vision--ECCV 2020: 16th European Conference, Glasgow, UK, August 23--28, 2020, Proceedings, Part II 16}, pages 402--419. Springer, 2020.

\bibitem[Tu et~al.(2024)Tu, Xing, Han, Cheng, Dai, Luo, and Wu]{tu2024stableanimator}
Shuyuan Tu, Zhen Xing, Xintong Han, Zhi-Qi Cheng, Qi Dai, Chong Luo, and Zuxuan Wu.
\newblock Stableanimator: High-quality identity-preserving human image animation.
\newblock \emph{arXiv preprint arXiv:2411.17697}, 2024.

\bibitem[Unterthiner et~al.(2018)Unterthiner, Van~Steenkiste, Kurach, Marinier, Michalski, and Gelly]{unterthiner2018towards}
Thomas Unterthiner, Sjoerd Van~Steenkiste, Karol Kurach, Raphael Marinier, Marcin Michalski, and Sylvain Gelly.
\newblock Towards accurate generative models of video: A new metric \& challenges.
\newblock \emph{arXiv preprint arXiv:1812.01717}, 2018.

\bibitem[Wang et~al.(2025)Wang, Ai, Wen, Mao, Xie, Chen, Yu, Zhao, Yang, Zeng, Wang, Zhang, Zhou, Wang, Chen, Zhu, Zhao, Yan, Huang, Feng, Zhang, Li, Wu, Chu, Feng, Zhang, Sun, Fang, Wang, Gui, Weng, Shen, Lin, Wang, Wang, Zhou, Wang, Shen, Yu, Shi, Huang, Xu, Kou, Lv, Li, Liu, Wang, Zhang, Huang, Li, Wu, Liu, Pan, Zheng, Hong, Shi, Feng, Jiang, Han, Wu, and Liu]{wan2025}
Ang Wang, Baole Ai, Bin Wen, Chaojie Mao, Chen-Wei Xie, Di Chen, Feiwu Yu, Haiming Zhao, Jianxiao Yang, Jianyuan Zeng, Jiayu Wang, Jingfeng Zhang, Jingren Zhou, Jinkai Wang, Jixuan Chen, Kai Zhu, Kang Zhao, Keyu Yan, Lianghua Huang, Mengyang Feng, Ningyi Zhang, Pandeng Li, Pingyu Wu, Ruihang Chu, Ruili Feng, Shiwei Zhang, Siyang Sun, Tao Fang, Tianxing Wang, Tianyi Gui, Tingyu Weng, Tong Shen, Wei Lin, Wei Wang, Wei Wang, Wenmeng Zhou, Wente Wang, Wenting Shen, Wenyuan Yu, Xianzhong Shi, Xiaoming Huang, Xin Xu, Yan Kou, Yangyu Lv, Yifei Li, Yijing Liu, Yiming Wang, Yingya Zhang, Yitong Huang, Yong Li, You Wu, Yu Liu, Yulin Pan, Yun Zheng, Yuntao Hong, Yupeng Shi, Yutong Feng, Zeyinzi Jiang, Zhen Han, Zhi-Fan Wu, and Ziyu Liu.
\newblock Wan: Open and advanced large-scale video generative models.
\newblock \emph{arXiv preprint arXiv:2503.20314}, 2025.

\bibitem[Wang et~al.(2024)Wang, Chen, Ma, Zhou, Huang, Wang, Yang, He, Yu, Yang, et~al.]{wang2024lavie}
Yaohui Wang, Xinyuan Chen, Xin Ma, Shangchen Zhou, Ziqi Huang, Yi Wang, Ceyuan Yang, Yinan He, Jiashuo Yu, Peiqing Yang, et~al.
\newblock Lavie: High-quality video generation with cascaded latent diffusion models.
\newblock \emph{International Journal of Computer Vision}, pages 1--20, 2024.

\bibitem[Wang et~al.(2004)Wang, Bovik, Sheikh, and Simoncelli]{wang2004image}
Zhou Wang, Alan~C Bovik, Hamid~R Sheikh, and Eero~P Simoncelli.
\newblock Image quality assessment: from error visibility to structural similarity.
\newblock \emph{IEEE transactions on image processing}, 13\penalty0 (4):\penalty0 600--612, 2004.

\bibitem[Xu et~al.(2024)Xu, Zou, Huang, Chen, Liu, Cheng, Shi, and Huang]{xu2024easyanimate}
Jiaqi Xu, Xinyi Zou, Kunzhe Huang, Yunkuo Chen, Bo Liu, MengLi Cheng, Xing Shi, and Jun Huang.
\newblock Easyanimate: A high-performance long video generation method based on transformer architecture.
\newblock \emph{arXiv preprint arXiv:2405.18991}, 2024.

\bibitem[Yang et~al.(2024)Yang, Teng, Zheng, Ding, Huang, Xu, Yang, Hong, Zhang, Feng, et~al.]{yang2024cogvideox}
Zhuoyi Yang, Jiayan Teng, Wendi Zheng, Ming Ding, Shiyu Huang, Jiazheng Xu, Yuanming Yang, Wenyi Hong, Xiaohan Zhang, Guanyu Feng, et~al.
\newblock Cogvideox: Text-to-video diffusion models with an expert transformer.
\newblock \emph{arXiv preprint arXiv:2408.06072}, 2024.

\bibitem[Zhang et~al.(2023)Zhang, Wang, Zhang, Zhao, Yuan, Qin, Wang, Zhao, and Zhou]{zhang2023i2vgen}
Shiwei Zhang, Jiayu Wang, Yingya Zhang, Kang Zhao, Hangjie Yuan, Zhiwu Qin, Xiang Wang, Deli Zhao, and Jingren Zhou.
\newblock I2vgen-xl: High-quality image-to-video synthesis via cascaded diffusion models.
\newblock \emph{arXiv preprint arXiv:2311.04145}, 2023.

\bibitem[Zhou et~al.(2021)Zhou, Sun, Wu, Loy, Wang, and Liu]{zhou2021pose}
Hang Zhou, Yasheng Sun, Wayne Wu, Chen~Change Loy, Xiaogang Wang, and Ziwei Liu.
\newblock Pose-controllable talking face generation by implicitly modularized audio-visual representation.
\newblock In \emph{Proceedings of the IEEE/CVF conference on computer vision and pattern recognition}, pages 4176--4186, 2021.

\end{thebibliography}
}

\end{document}